\documentstyle[eaclap]{article}

\title{\vspace{-0.5in}Computational dialectology in Irish Gaelic}
\author{Brett Kessler\thanks{I thank Martin Kay, Paul Kiparsky, Tom Wasow,
and a reviewer for helpful comments.}\\
Department of Linguistics\\
Stanford University \\
Stanford CA 94305-2150 USA\\
kessler@csli.stanford.edu\\}

\begin{document}

\maketitle
\bibliographystyle{acl}

\vspace{-0.5in}
\begin{abstract}

Dialect groupings can be discovered objectively and automatically by
cluster analysis of phonetic transcriptions such as those found in a
linguistic atlas.  The first step in the analysis, the computation of
linguistic distance between each pair of sites, can be computed as
Levenshtein distance between phonetic strings.  This correlates
closely with the much more laborious technique of determining and
counting isoglosses, and is more accurate than the more familiar
metric of computing Hamming distance based on whether vocabulary
entries match.  In the actual clustering step, traditional
agglomerative clustering works better than the top-down technique of
partitioning around medoids.  When agglomerative clustering of
phonetic string comparison distances is applied to Gaelic,
reasonable dialect boundaries are obtained, corresponding to national
and (within Ireland) provincial boundaries.

\end{abstract}

\section{Introduction}

Defining dialects is one of the first tasks that linguists need to pursue when
approaching a language.  Knowing the dialect areas helps one allocate
resources in language research and has implications for language learners,
publishers, broadcasters, educators, and language planners.  Unfortunately,
dialect definition can be a time-consuming and ill-defined process.  The
traditional approach has been to plot isoglosses, delineating regions where
the same word is used for the same concept, or perhaps the same pronunciation
for the same phoneme.  But isoglosses are frustrating.  The first problem, as
Gaston Paris noted (apud Durand, 1889:49), is that isoglosses rarely coincide.
At best, isoglosses for different features approach each other, forming vague
bundles; at worst, isoglosses may cut across each other, describing completely
contradictory binary divisions of the dialect area.  That is, language may
vary geographically in many dimensions, but the requirements we usually impose
require that a specific site be placed in a unique dialect.  Traditional
dialectological methodology gives little guidance as to how to perform such
reduction to one dimension.

A second problem is that many isoglosses do not neatly bisect the language
area.  Often variants do not neatly line up on two sides of a line, but are
intermixed haphazardly.  More importantly, for some sites information may be
lacking, or the question is simply not applicable.  When comparing how various
sites pronounce the first consonant of a particular word, it is meaningless to
ask that question if the site does not use that word.  So the isogloss is
incomplete and cannot be meaningfully compared with isoglosses based on
different sets of sites.

The third problem is that most languages have dialect continua, such that the
speech of one community differs little from the speech of its neighbours.
Even though the cumulative effects of such differences may be great when one
considers the ends of the continua (such as southern Italian versus northern
French), still it seems arbitrary to draw major dialect boundaries between two
villages with very similar speech patterns.  Such conundrums led Paris and
others to conclude that the dialect boundary, and therefore the very notion of
dialect, is an ill-defined concept.

More recently, the field of dialectometry, as introduced by S\'eguy
(1971, 1973), has addressed these issues by developing several
techniques for summarizing and presenting variation along multiple
dimensions.  They replace isoglosses with a distance matrix, which
compares each site directly with all other sites, ultimately yielding
a single figure that measures the linguistic distances between each
pair of sites.  There is however no firm agreement on just how to
compute the distance matrices.  S\'eguy's earliest work (1971) was
based on lexical correspondences: sites differed in the extent to
which they used different words for the same concept.  S\'eguy (1973),
Philps (1987), and Durand (1989) use some combination of lexical,
phonological, and morphological data.  Babitch (1988) described the
dialectal distances in Acadian villages by the degree to which their
fishing terminology varied.  Babitch and Lebrun (1989) did a similar
analysis based on the varying pronunciation of /r/.  Elsie (1986)
grouped the Gaelic dialects on the basis of whether the vocabulary
matched.  Ebobisse (1989) grouped the Sawabantu languages of Cameroon
by whether phonological correspondences in matching vocabulary items
were complete, partial, or lacking.  There seems to be a certain bias
in favour of working with lexical correspondences, which is
understandable, since deciding whether two sites use the same word for
the same concept is perhaps one of the easiest linguistic judgements
to make.  The need to figure out such systems as the comparative
phonology of various linguistic sites can be very time-consuming and
fraught with arbitrary choices.

Not all dialectometrists agree on the wisdom of delineating dialect
areas.  S\'eguy (1973:18) insisted that the concept of dialect
boundaries was meaningless, and his emphasis on the gradience of
language similarity has been widely maintained.  But those who do look
for firm dialect affiliations (such as Babitch and Ebobisse) use
bottom-up agglomerative techniques.  The two linguistically closest
sites are grouped into one dialect, and thenceforth treated as a unit.
The process continues recursively until all sites are grouped into one
superdialect embracing the entire language area under consideration.
This yields a binary tree.  But Kaufman and Rousseeuw (1990:44)
suggest that when the emphasis in a clustering problem is on the
top-level clusters---here, finding the two main dialects---then such
bottom-up methods, which can potentially introduce error at each of
several steps, are less reliable than top-down partitioning methods.
Perhaps past researchers have used inferior bottom-up techniques
simply because the necessary algorithms are computationally more
tractable.  Comparing all possible pairs of sites is a O($N^2$)
problem,\footnote{The overall algorithm is O($N^3$) since for each new
group one must compute the distances between it and each of the other
sites or groups.} whereas considering all possible two-way partitions
of the dialect area is O(2$^N$).

The current state of dialectometry thus presents two main questions which
constitute the methodological focus of this paper.  The first deals with
distance matrices.  Is there a way to build accurate distance matrices that
minimize editorial decisions without discarding relevant data?  My research
suggests that this may be done by using string distances computed directly on
phonetic transcriptions, and that this is better than restricting the study to
lexical comparisons.  The second deals with clustering.  Do bottom-up or
top-down techniques work best?  My conclusion is that the traditional
bottom-up technique works better than a typical top-down method.  These
conclusions are based partly on an analysis of the mathematical properties
of the clusters themselves, partly on how well they correlate with analyses
based on more traditional isogloss techniques, and partly on how well they
compare with previously-published descriptions of dialects in a specific
language, Irish Gaelic.

At one time the Gaelic language group was spoken throughout Ireland,
from where it spread to the Isle of Man and to much of Scotland.
Currently fully native use of Gaelic is limited to a few discontiguous
areas in the westernmost reaches of Ireland and Scotland.  In the case
of Ireland, everyone agrees that Gaelic is nowadays found in three
main dialects: that of Ulster, that of Connacht, and that of Munster
(\'O~Siadhail, 1989).  But several questions are raised that are less
easily answered.  Do the three provinces separate out so neatly for
intrinsic linguistic reasons, or simply because their speakers have
become so widely separated from each other geographically as speakers
in intervening areas have adopted English?  Does the language of
Connacht naturally group with that of Ulster or with that of Munster?
And looking beyond Ireland, many have commented that the language of
Ulster in general is similar to that of Scotland.  Are Irish, Manx,
and Scottish Gaelic considered three separate languages for intrinsic
linguistic reasons, or because they are spoken in different countries?
To a large extent, dialectologists have found these questions
difficult to answer because they accepted Paris's conundrum.  For
\'O~Siadhail, the ultimate scientific justification in adopting the
three-dialect account is the fact that the Gaeltacht (Irish-speaking
territory) is so fragmented nowadays that it no longer forms a
continuum.  \'O~Cu{\'\i}v (1951:4--49) felt that there can be no
dialect boundaries because transitions are gradual.  Elsie (1986:240)
considers a dialect to be an area where all communities are
linguistically more similar to each other than any community is to any
site outside the dialect.  Such notions provide a very firm, absolute
notion of dialecthood: a set of communities either constitutes a
dialect area, or it does not.  But as the dialectometrists have shown,
other notions of clustering are equally scientific and may more
accurately correspond to intuitive notions of what it means to be a
dialect.

\section{Data}

The data for my study were taken from Wagner 1958.  Wagner
administered a questionnaire to native speakers of Irish Gaelic in 86
sites.\footnote{The atlas also maps for Kilkenny some information
gathered from another source.}  Most of the informants were over
seventy years old and had not spoken Irish since their youth.  The
atlas is therefore an approximate reconstruction of the linguistic
landscape of the turn of the century, when the Gaeltacht was more
continuous.  Wagner also presents material from the Isle of Man and
seven sites in Scotland.  The mapped entries are presented in a very
narrow phonetic transcription based on the International Phonetic
Alphabet.

Volume 1 of Wagner 1958 consists of 300 maps, plotting about 370
concepts.  I used the first 51 concepts, or about 4500 different string
tokens, as part of an ongoing project to enter all of the atlas into
machine readable format.  These 51 concepts were represented by 312
different Gaelic words or phrases, whose stems derived from 171
different etymons.

\section{Methodology and results}

\subsection{Distance matrices}

To form a baseline for comparison, I analysed the distribution of each
of the 51 plotted concepts, finding a total of 3,337 features by which
two or more sites differed.  For example, for the concept `sell', I
identified two sets, one using the word {\em d{\'\i}ol} (most sites in
Ireland), and one using the word {\em creic} (Rathlin Island, the Isle
of Man, and Scotland).  The dialects partitioned in a different way
according to how much stress they placed on the verb relative to the
pronoun in `I sold' (even stress in Dunlewy and four of the Scottish
sites, else extra stress on the verb).  Not all partitions covered the
entire map. In this example, only sites that used the word {\em
d{\'\i}ol} were compared on the basis of whether a schwa developed in
the sequence [i:l].  In some cases the divisions were more than
two-way: for example, Wagner distinguishes whether the final consonant
in {\em creic} is unpalatalized, palatalized, or slightly palatalized.
Distance between sites was determined by counting 0 whenever two sites
were in the same set and 1 whenever two sites were in contrasting
sets, then taking the mean.  This baseline approach corresponds
formally to determining distance by the number of isoglosses that
separate sites, which is in principle the traditional technique.

This baseline was compared to several other approaches.  The {\em etymon
identity} metric averaged the number of times the sites agreed in using words
whose stem had the same ultimate derivation.  For example, the dialects
differed as to whether they used some form of {\em bull-} or {\em damh-} for
the word `bullock'.  Etymon identity is one of the more common approaches in
dialectometry; Elsie for example used it in his study of the Gaelic dialects
(1986).  Closely related is the idea of {\em word identity}, where the words
are not counted the same unless all of their morphemes agree.  In this
analysis, sites that used some form of the word {\em bull\'an}, with the
suffix {\em -\'an}, were distinguished from those using the suffix {\em
-\'og}.

\newcommand{\smallcap}[1]{{\small\sc #1}}
\newcommand{\bigA}{\smallcap{a}}
\newcommand{\bigL}{\smallcap{l}}

Another set of approaches for computing distance was based on the
phonetics.  This computed the Levenshtein distance between phonetic
strings.  The Levenshtein distance is the cost of the least expensive
set of insertions, deletions, or substitutions that would be needed to
transform one string into the other (Sankoff and Kruskal, 1983).  The
simplest technique used was {\em phone string comparison}.  In this
approach, all operations cost 1 unit.  Thus in comparing the forms
[{\bigA}{\bigL}:i] and [a{\bigL}i] for {\em eallaigh} `cattle', the
(minimal) distance was 2, for the substitutions [a]/[\bigA] and
[\bigL:]/[\bigL].  (For this measure, diacritics such as the length
mark `:' were counted as part of the letter, and different diacritics
were adjudged to make for different letters.)  A pair of unrelated
words like [\bigA\bigL:i] and [khruh] (for {\em crodh}, another word
for `cattle') would get a much larger score, 5.

In the above technique, very small phonetic differences, such as that
between a moderately palatalized and a very palatalized [t], count the
same as major differences, such as that between a [t] and an [e].  It
would seem to be more accurate to assign a greater distance to
substitutions involving greater phonetic distinctions.  Unfortunately
I know of no comprehensive study on the differences between phones, at
least not for all 277 contrasts made by Wagner.  Instead I
distinguished them on the basis of twelve phonetic features that
systematically account for all of the distinctions in Wagner's
inventory: nasality, stricture, laterality, articulator, glottis,
place, palatalization, rounding, length, height, strength, and
syllabicity.  The features were given discrete ordinal values scaled
between 0 and 1, the exact values being arbitrary.  For example, {\em
place} took on the values {\em glottal=0, uvular=0.1, postvelar=0.2,
velar=0.3, prevelar=0.4, palatal=0.5, alveolar=0.7, dental=0.8,} {\em
and labial=1.} The distance between any two phones was judged to be
the difference between the feature values, averaged across all twelve
features. These distances were used instead of uniform 1-unit costs in
computing Levenshtein distance.  The resulting metric was called {\em
feature string comparison}.

It could be argued that it is meaningless to compare the phonetics of
different words, as in the case of {\em eallaigh} vs.\ {\em crodh}
mentioned above.  Therefore the feature string comparison was also
computed only for pairs of citations that used the same word, so that
[\bigA\bigL:i] vs.\ [a{\bigL}i] would be compared, but [\bigA\bigL:i]
vs.\ [khruh] would be ignored.  The different approaches are called
{\em all-word} vs.\ {\em same-word} feature string comparisons.

All of these distance matrices were compared with the isogloss matrix,
to see which of them gives results closest to that base method.  I
used two different methods of comparison, Pearson's $\rho$ computed
between all corresponding cells in the two matrices, and \[K_c =
Average(sign((X_{ij} - X_{ik})(Y_{ij} - Y_{ik}))\] which is a derivative of
Kendall's $\tau$ that Dietz (1983) empirically found particularly
accurate as a test statistic for comparing distance
matrices.\footnote{That is, for each site $i$, one considers all other
pairs of sites, $j$ and $k$, and asks whether the linguistic
difference between $i$ and $j$ is greater or less than that between
$i$ and $k$.  One counts 1 if the answer is the same for both distance
matrices, $-1$ if it is different.  $K_c$ is the average of these
numbers.}  Table~\ref{Matrix-comparison} shows that the two measures
give parallel results.  More importantly, it shows that the approaches
based on string comparisons of the phonetic transcriptions correlate
more highly with the isogloss approach than do the word or etymon
identity measures.  Furthermore, comparing whole phone letters works
better than the more sophisticated technique of comparing features,
and restricting comparison to pairs based on the same words does not
make the latter any better.

Of course, I do not expect that this technique using flat 1-unit costs
will prove superior to all methods that are more sensitive to phonetic
details.  Feature comparison may work better if features were weighted
differentially, or if the numeric values they assume were assigned
less arbitrarily, or if the Manhattan-style distance computation were
replaced by some formula that did not assume that the features are
independent of each other.  An ideal comparison would be based on data
telling how likely it is for the one phone to turn into the other in the
course of normal language change. In the method described here, [s] is
adjudged closer to [g] than to [h].  But [s] often changes into [h] in
the world's languages, and so the pair should have a small distance;
whereas the change of [s] to [g] has never occurred to my knowledge,
and so should have a very large distance.  The unfortunate fact that
such ideal data are lacking is compensated for by the fact that the
inexpensive phone-string comparison employed in this study performs
quite well.

\begin{table}
\caption{\label{Matrix-comparison}Correlation of distance matrices to the
isogloss distance matrix}
\begin{tabular}{lll}
& $\rho$ & $K_c$ \\
Phone string comparison         &     .95  &   .76 \\
Feature string comparison \\
{---}{---} all-word       &     .92  &   .70 \\
{---}{---} same-word       &     .91 & .69 \\
Etymon identity                 &     .85  &   .61 \\
Word identity                   &     .84  &   .63 \\
\end{tabular}
\end{table}

\subsection{Clustering techniques}

The traditional agglomerative technique for clustering has been
described above.  There is some variation in how the distance between
two clusters is measured.  For this study I used the average distance
between all pairs of elements that are in different clusters.  I
compared agglomeration to a top-down method that Kaufman and Rousseeuw
(1990) call {\em partitioning around medoids}. This model reduces the
O($2^N$) intractability of top-down approaches discussed above by
dramatically reducing the number of binary partitions that are
considered.  Here one seeks to divide the sites into two groups by
finding the two representative sites (the medoids) around which all
the other sites cluster in such a way as to give the least average
distance between the sites and their representatives. This is
therefore a O($N^3$) algorithm, comparable in efficiency to
agglomeration.  The process was repeated recursively on each dialect.

One way of measuring how well a binary clustering technique works for dialect
grouping is to compare for each site $i$ its average dissimilarity from the
other sites in the same dialect, $a(i)$, with its average dissimilarity from
the sites in the other dialect, $b(i)$.  Kaufman and Rousseeuw (1990:83--86)
define the statistic $s(i)$ to be \(1 - a(i)/b(i)\) if $a(i)$ is less than
$b(i)$, otherwise \(b(i)/a(i) - 1\).  The statistic thus ranges from 1
(perfect fit) to $-1$ (site $i$ would perfectly fit in the other group).
Plotting this statistic gives a {\em silhouette} by which the eye can judge
how well classified each site is.  Averaging this statistic across all sites
gives an idea of how felicitous the overall clustering is, $\bar{s}$.

Figures~\ref{partition-silhouette1}--\ref{partition-silhouette2}
present the silhouette for clustering the isogloss distance matrix by
partitioning.  This analysis produces a large dialect which groups
together the sites in Munster, Scotland, the Isle of Man, and almost
all sites in Connacht, as well as Rathlin Island in Ulster; and
another which groups together all the other sites in Ulster, as well
as County Cavan in Connacht.  Although the Ulster group is
fairly tight, with an $\bar{s}$ of 0.41, the other group has a more
anemic $\bar{s}$ of 0.25, with the sites outside of Munster and
Southern Connacht being indifferently classified.  The weighted
$\bar{s}$ for both groups comes out at 0.29.  By comparison,
Figures~\ref{agglomeration-silhouette1}--\ref{agglomeration-silhouette2}
show what happens when the traditional agglomerative technique is
used.  The dialects of Scotland and the Isle of Man form a cluster
with a great deal of internal diversity (\(\bar{s} = 0.12\)), and all
the sites in Ireland form another cluster averaging \(\bar{s} =
0.37\), with only Rathlin Island being indifferently classified.  The
weighted average is 0.35, which is superior to that of the
partitioning technique.

\begin{figure}
\tiny
\begin{tabular}{ll}
{*}*** &  Kilkenny, Kilkenny, Leinster, Ireland \\
{*}*** &  Lough Attorick, Galway, Connacht, Ireland \\
{*}** &   Doolin, Clare, Munster, Ireland \\
{*}** &   Fanore, Clare, Munster, Ireland \\
{*}** &   Clear Island, Cork, Munster, Ireland \\
{*}** &   Skibbereen, Cork, Munster, Ireland \\
{*}** &   Kinvra, Galway, Connacht, Ireland \\
{*}** &   Coomhola, Cork, Munster, Ireland \\
{*}** &   Cloghane, Kerry, Munster, Ireland \\
{*}** &   Kilgarvan, Kerry, Munster, Ireland \\
{*}** &   Waterville, Kerry, Munster, Ireland \\
{*}** &   Killorglin, Kerry, Munster, Ireland \\
{*}** &   Glandore, Cork, Munster, Ireland \\
{*}** &   Carraroe, Galway, Connacht, Ireland \\
{*}** &   Dursey Sound, Cork, Munster, Ireland \\
{*}** &   Careeny, Galway, Connacht, Ireland \\
{*}** &   Ballymacoda, Cork, Munster, Ireland \\
{*}** &   Newbridge, Galway, Connacht, Ireland \\
{*}** &   Kilsheelan, Waterford, Munster, Ireland \\
{*}** &   Conakilty, Cork, Munster, Ireland \\
{*}** &   Craughwell, Galway, Connacht, Ireland \\
{*}** &   Coolea, Cork, Munster, Ireland \\
{*}** &   Rosmuck, Galway, Connacht, Ireland \\
{*}** &   Glenflesk, Kerry, Munster, Ireland \\
{*}** &   Mount Melleray, Waterford, Munster, Ireland \\
{*}** &   Cornamona, Galway, Connacht, Ireland \\
{*}** &   Dunquin, Kerry, Munster, Ireland \\
{*}** &   Tourmakeady, Mayo, Connacht, Ireland \\
{*}** &   Ring, Waterford, Munster, Ireland \\
{*}** &   Laughanbeg, Galway, Connacht, Ireland \\
{*}** &   Emlaghmore, Galway, Connacht, Ireland \\
{*}* &    Lauragh, Kerry, Munster, Ireland \\
{*}* &    Kilbaha, Clare, Munster, Ireland \\
{*}* &    Moycullen, Galway, Connacht, Ireland \\
{*}* &    Sliabh gCua, Waterford, Munster, Ireland \\
{*}* &    Kilmovee, Mayo, Connacht, Ireland \\
{*}* &    Goatenbridge, Tipperary, Munster, Ireland \\
{*}* &    Colmanstown, Galway, Connacht, Ireland \\
{*}* &    Inisheer, Galway, Connacht, Ireland \\
{*}* &    Glentrasna, Galway, Connacht, Ireland \\
{*}* &    Letterfrack, Galway, Connacht, Ireland \\
{*}* &    Angliham, Galway, Connacht, Ireland \\
{*}* &    Annaghdown, Galway, Connacht, Ireland \\
{*}* &    Lough Nafooey, Galway, Connacht, Ireland \\
{*}* &    Sonnagh, Galway, Connacht, Ireland \\
{*}* &    C\'arna, Galway, Connacht, Ireland \\
{*}* &    Louisburgh, Mayo, Connacht, Ireland \\
{*}* &    Inishmaan, Galway, Connacht, Ireland \\
{*}* &    Camderry, Galway, Connacht, Ireland \\
{*}* &    Ceathr\'u an Tairbh, Roscommon, Connacht, Ire. \\
{*}* &    Carnmore, Galway, Connacht, Ireland \\
{*}* &    Ballycastle, Mayo, Connacht, Ireland \\
{*}* &    Cashel, Galway, Connacht, Ireland \\
{*}* &    Tobercurry, Sligo, Connacht, Ireland \\
{*}* &    Ballyglunin, Galway, Connacht, Ireland \\
{*}* &    Aclare, Sligo, Connacht, Ireland \\
{*}* &    Belderg, Mayo, Connacht, Ireland \\
{*} &     Dohooma, Mayo, Connacht, Ireland \\
{*} &     Portacloy, Mayo, Connacht, Ireland \\
{*} &     Blacksod, Mayo, Connacht, Ireland \\
{*} &     Kintyre, Argyll, Scotland \\
   &    Isle of Man \\
   &    Slievenakilla, Leitrim, Connacht, Ireland \\
    &   Inveraray, Argyll, Scotland \\
   &    Arran, Bute, Scotland \\
   &    Curraun Peninsula, Mayo, Connacht, Ireland \\
   &    Achill, Mayo, Connacht, Ireland \\
   &    Lochalsh, Ross and Cromarty, Scotland \\
   &    Assynt, Sutherland, Scotland \\
    &   Carloway, Lewis, Ross and Cromarty, Scotland \\
    &   Benbecula, Inverness, Scotland \\
    &   Ballyconnell, Sligo, Connacht, Ireland \\
    &   Rathlin Island, Antrim, Ulster, Ireland \\
\end{tabular}
\caption{\label{partition-silhouette1}Silhouette for the first top-level
binary dialect grouping computed on the isogloss distance matrix via
partitioning.  Stars represent relative $s(i)$.  Locations in Ireland
are cited by locality, county, province, and country.}
\end{figure}

\begin{figure}
\tiny
\begin{tabular}{ll}
{*}**** & Kildarragh, Donegal, Ulster, Ireland \\
{*}*** &  Creeslough, Donegal, Ulster, Ireland \\
{*}*** &  Glenvar, Donegal, Ulster, Ireland \\
{*}*** &  Loughanure, Donegal, Ulster, Ireland \\
{*}*** &  Lettermacaward, Donegal, Ulster, Ireland \\
{*}*** &  Beflaght, Donegal, Ulster, Ireland \\
{*}*** &  Kingarroo, Donegal, Ulster, Ireland \\
{*}*** &  Croaghs, Donegal, Ulster, Ireland \\
{*}*** &  Aranmore, Donegal, Ulster, Ireland \\
{*}*** &  Gortahork, Donegal, Ulster, Ireland \\
{*}*** &  Downings, Donegal, Ulster, Ireland \\
{*}*** &  Tory Island, Donegal, Ulster, Ireland \\
{*}*** &  Dunlewy, Donegal, Ulster, Ireland \\
{*}** &   Rannafast, Donegal, Ulster, Ireland \\
{*}** &   Meenacharvy, Donegal, Ulster, Ireland \\
{*}** &   Teelin, Donegal, Ulster, Ireland \\
{*}** &   Ardara, Donegal, Ulster, Ireland \\
{*}** &   Ballyhooriskey, Donegal, Ulster, Ireland \\
{*}** &   Creggan, Tyrone, Ulster, Ireland \\
{*}** &   Clonmany, Donegal, Ulster, Ireland \\
{*}* &    Omeath, Louth, Ulster, Ireland \\
{*} &     Glangevlin, Cavan, Connacht, Ireland \\
\end{tabular}
\caption{\label{partition-silhouette2}Silhouette for the second
dialect grouping computed on the isogloss distance matrix via
partitioning.}
\end{figure}

\begin{figure}
\tiny
\begin{tabular}{ll}
{*}                  & Carloway, Lewis, Ross and Cromarty, Scotland \\
{*}                  & Benbecula, Inverness, Scotland \\
{*}                  & Assynt, Sutherland, Scotland \\
{*}                  & Inveraray, Argyll, Scotland \\
{*}                  & Lochalsh, Ross and Cromarty, Scotland \\
{*}                  & Kintyre, Argyll, Scotland \\
{*}                  & Arran, Bute, Scotland \\
\                   & Isle of Man \\
\end{tabular}
\caption{\label{agglomeration-silhouette1}Silhouette for isogloss dialect
grouping using agglomerative clustering, first group.}
\end{figure}

\begin{figure}
\tiny
\begin{tabular}{ll}
{*}***               & Colmanstown, Galway, Connacht, Ireland \\
{*}***               & Moycullen, Galway, Connacht, Ireland \\
{*}***               & Ceathr\'u an Tairbh, Roscommon, Connacht, Ire. \\
{*}***               & Ballyconnell, Sligo, Connacht, Ireland \\
{*}***               & Carnmore, Galway, Connacht, Ireland \\
{*}***               & Annaghdown, Galway, Connacht, Ireland \\
{*}***               & Lough Nafooey, Galway, Connacht, Ireland \\
{*}***               & Glentrasna, Galway, Connacht, Ireland \\
{*}***               & Ballycastle, Mayo, Connacht, Ireland \\
{*}***               & Carraroe, Galway, Connacht, Ireland \\
{*}***               & Cornamona, Galway, Connacht, Ireland \\
{*}***               & Aclare, Sligo, Connacht, Ireland \\
{*}***               & Emlaghmore, Galway, Connacht, Ireland \\
{*}***               & Dohooma, Mayo, Connacht, Ireland \\
{*}***               & Ballyglunin, Galway, Connacht, Ireland \\
{*}***               & Sonnagh, Galway, Connacht, Ireland \\
{*}***               & Cashel, Galway, Connacht, Ireland \\
{*}***               & Curraun Peninsula, Mayo, Connacht, Ireland \\
{*}***               & Craughwell, Galway, Connacht, Ireland \\
{*}***               & Belderg, Mayo, Connacht, Ireland \\
{*}***               & Portacloy, Mayo, Connacht, Ireland \\
{*}***               & Angliham, Galway, Connacht, Ireland \\
{*}***               & Blacksod, Mayo, Connacht, Ireland \\
{*}***               & Laughanbeg, Galway, Connacht, Ireland \\
{*}***               & Kinvra, Galway, Connacht, Ireland \\
{*}***               & Coomhola, Cork, Munster, Ireland \\
{*}***               & Camderry, Galway, Connacht, Ireland \\
{*}***               & Cloghane, Kerry, Munster, Ireland \\
{*}***               & Rosmuck, Galway, Connacht, Ireland \\
{*}***               & Newbridge, Galway, Connacht, Ireland \\
{*}***               & Lough Attorick, Galway, Connacht, Ireland \\
{*}***               & Kilmovee, Mayo, Connacht, Ireland \\
{*}***               & Achill, Mayo, Connacht, Ireland \\
{*}**                & Dursey Sound, Cork, Munster, Ireland \\
{*}**                & Tourmakeady, Mayo, Connacht, Ireland \\
{*}**                & Letterfrack, Galway, Connacht, Ireland \\
{*}**                & Clear Island, Cork, Munster, Ireland \\
{*}**                & Skibbereen, Cork, Munster, Ireland \\
{*}**                & Glandore, Cork, Munster, Ireland \\
{*}**                & Tobercurry, Sligo, Connacht, Ireland \\
{*}**                & Glenflesk, Kerry, Munster, Ireland \\
{*}**                & Fanore, Clare, Munster, Ireland \\
{*}**                & Careeny, Galway, Connacht, Ireland \\
{*}**                & Doolin, Clare, Munster, Ireland \\
{*}**                & Killorglin, Kerry, Munster, Ireland \\
{*}**                & Dunquin, Kerry, Munster, Ireland \\
{*}**                & Louisburgh, Mayo, Connacht, Ireland \\
{*}**                & Kilsheelan, Waterford, Munster, Ireland \\
{*}**                & Waterville, Kerry, Munster, Ireland \\
{*}**                & Kilgarvan, Kerry, Munster, Ireland \\
{*}**                & C\'arna, Galway, Connacht, Ireland \\
{*}**                & Ballymacoda, Cork, Munster, Ireland \\
{*}**                & Conakilty, Cork, Munster, Ireland \\
{*}**                & Kilbaha, Clare, Munster, Ireland \\
{*}**                & Coolea, Cork, Munster, Ireland \\
{*}**                & Lauragh, Kerry, Munster, Ireland \\
{*}**                & Glangevlin, Cavan, Connacht, Ireland \\
{*}**                & Sliabh gCua, Waterford, Munster, Ireland \\
{*}**                & Mount Melleray, Waterford, Munster, Ireland \\
{*}**                & Kingarroo, Donegal, Ulster, Ireland \\
{*}**                & Goatenbridge, Tipperary, Munster, Ireland \\
{*}**                & Downings, Donegal, Ulster, Ireland \\
{*}**                & Croaghs, Donegal, Ulster, Ireland \\
{*}**                & Inishmaan, Galway, Connacht, Ireland \\
{*}**                & Glenvar, Donegal, Ulster, Ireland \\
{*}**                & Ring, Waterford, Munster, Ireland \\
{*}**                & Lettermacaward, Donegal, Ulster, Ireland \\
{*}**                & Kildarragh, Donegal, Ulster, Ireland \\
{*}**                & Inisheer, Galway, Connacht, Ireland \\
{*}**                & Creeslough, Donegal, Ulster, Ireland \\
{*}**                & Gortahork, Donegal, Ulster, Ireland \\
{*}**                & Beflaght, Donegal, Ulster, Ireland \\
{*}**                & Kilkenny, Kilkenny, Leinster, Ireland \\
{*}*                 & Ardara, Donegal, Ulster, Ireland \\
{*}*                 & Rannafast, Donegal, Ulster, Ireland \\
{*}*                 & Aranmore, Donegal, Ulster, Ireland \\
{*}*                 & Dunlewy, Donegal, Ulster, Ireland \\
{*}*                 & Meenacharvy, Donegal, Ulster, Ireland \\
{*}*                 & Loughanure, Donegal, Ulster, Ireland \\
{*}*                 & Slievenakilla, Leitrim, Connacht, Ireland \\
{*}*                 & Ballyhooriskey, Donegal, Ulster, Ireland \\
{*}*                 & Omeath, Louth, Ulster, Ireland \\
{*}*                 & Creggan, Tyrone, Ulster, Ireland \\
{*}*                 & Clonmany, Donegal, Ulster, Ireland \\
{*}*                 & Teelin, Donegal, Ulster, Ireland \\
{*}*                 & Tory Island, Donegal, Ulster, Ireland \\
                   & Rathlin Island, Antrim, Ulster, Ireland \\
\end{tabular}
\caption{\label{agglomeration-silhouette2}Silhouette by agglomeration,
Irish group.}
\end{figure}

The same comparative results obtain for almost all of the distance
measuring techniques.  Table~\ref{compare-clustering} shows that the
$\bar{s}$ for the first binary split is usually appreciably higher for
agglomeration than it is for partitioning.  This result suggests not
any inferiority of top-down techniques in general---applying the
$\bar{s}$ statistic to all binary partitions would by definition
reveal the optimal split---nor of partitioning around medoids in
general.  Rather, it appears that the assumption behind this
partitioning heuristic, that a site will be closer to the medoid of
its own group than to the medoid of the other group, often fails to
hold true in dialectology.  The lack of clean breaks between dialects
and the fact that dialects of the same language may vary greatly in
diameter (i.e., maximal intragroup distances) both mean that the
assumption will often be invalid.

\begin{table}
\caption{\label{compare-clustering}Statistic $\bar{s}$ for the top-level binary
dialect division, comparing partitioning around medoids and agglomeration for
the different distance matrices.}
\begin{tabular}{lcc}
            & Part. & Aggl. \\
Isoglosses &    0.287 &        0.345 \\
Phone string comparison & 0.185 & 0.322 \\
Feature string comparison \\
{---}{---} all-word       & 0.252 & 0.353 \\
{---}{---} same-word       & 0.219 & 0.401 \\
Etymon identity & 0.363 & 0.478 \\
Word identity & 0.370 & 0.309 \\
\end{tabular}
\end{table}

\subsection{Gaelic dialects}

Since agglomeration is the better clustering technique, the best
dialect analysis should be obtained by agglomerating the isogloss
matrix.  The best automated approximation should be agglomerating the
distance matrix computed by phonetic string comparison, and indeed the
top-level topologies produced by both techniques are virtually
identical.  Both group into one loosely-connected entity all the sites
in Scotland, and into another all the sites in Ireland.  The isogloss
approach groups Manx as a cousin of the Scottish dialects, and the
phonetic approach makes it a cousin of the Irish dialects, but in both
cases the $s$ of Manx is very small (less than 0.06), making it
essentially intermediate between the two groups.  Thus the popular
notion that Scottish, Irish and Manx Gaelic are distinct entities is
well supported by the analyses.  Both analyses group Rathlin Island
very weakly with the rest of Irish, but the $s$ for Rathlin is so low
(less than 0.09) that its grouping too is essentially arbitrary.  This
aligns with the fact that authorities disagree as to whether it is a
dialect of Irish (as does Wagner) or of Scottish (O'Rahilly 1932:191).
Except for Rathlin Island, both methods group the Irish sites into one
group containing all the sites in Ulster, and another, Southern,
group, which itself breaks into a group containing all the sites in
Connacht and one containing all the sites in Munster.\footnote{The one
site in Co.\ Cavan is intermediate between the Ulster and the Southern
group.  Wagner also gives two sites in Leinster.  The more southern
site, in Kilkenny, groups with the Southern group, and the more
northern site, in Co.\ Louth, groups with Ulster, and indeed the
county used to be considered part of that province.}  Both methods
agree on how the 87 sites are distributed among these dialects.  This
three-way division accords with the universal perception that Ulster,
Connacht and Munster form the three major dialect groups.  The special
status of Ulster contradicts the position of O'Rahilly (1932:18) that
Connacht groups with Ulster to form a Northern dialect over against
Munster.  But it agrees with Elsie's finding (1986:255) that that
province is lexicostatistically more remote from Connacht and Munster
than those two are from each other.  Furthermore, Hindley reports
(1990:63) that speakers of other dialects often switch off radio
broadcasts in Ulster Irish, `which is very distinctive'.

Thus the classification of the major Gaelic dialects gives the same
general results by both distance metrics, if one discounts Manx and
Rathlin Island Gaelic, which are flagged as indifferent in both
schemes.  It is encouraging that the resultant dialect areas are
continuous, align with traditional provincial boundaries, and agree
with commonly accepted taxonomies.  However, dialect groupings at
narrower levels, such as the exact subgrouping of the major provincial
dialects, are at this point unstable.  This is no doubt to be
explained by the fairly small number of mapped concepts on which the
distance metrics are based (51).\footnote{S\'eguy (1973) cites
empirical research suggesting that general dialectometry requires
about a hundred concepts.}  As language differences get smaller, one
expects that more data will be required in order to elucidate them.

\section{Conclusions}

This experiment shows that an automatic procedure can reliably group a
language into its dialect areas, starting from nothing more than phonetic
transcriptions as commonly found in linguistic surveys.  Accurate distance
matrices, corresponding highly to those obtained by the tedious uncovering of
thousands of isoglosses, can be obtained by averaging the Levenshtein distance
between phonetic strings, weighting equally all insertion, deletion, and
substitution operations on the constituent phones.  This turns out to be quite
a bit more precise than the common technique of measuring distances by judging
etymon identity, and requires even less editorial work.  That phonetic
comparison is more precise is not particularly surprising, since etymon
identity ignores a wealth of phonetic, phonological, and morphological data,
whereas comparing phones has the side effect of also counting higher-level
variation: if words differ in morphemes, their phonetic difference is going to
be high.  As for clustering the sites into dialect areas, the familiar
bottom-up
agglomeration method proves superior to top-down partitioning around medoids.

Of course simply knowing the dialect areas is not the last word in
dialectology.  There remain such essential problems as identifying the
differing linguistic structures that characterize the dialects, and discovering
their history.  But all of these tasks will be greatly facilitated by a quick
and accurate grouping of the dialects.

\end{document}